\documentclass[sigconf, screen]{acmart}
\pdfoutput=1
\acmConference[AIR'23]{Advances in Robotics}{July,
  2023}{IIT Ropar}
\acmPrice{}
\acmISBN{}

\graphicspath{ {samples/images/} }
\usepackage[inkscapearea=page]{svg}
\usepackage{adjustbox}
\usepackage{pdfpages}
\usepackage{subfig}

\makeatletter
\setcopyright{none}
\renewcommand\@formatdoi[1]{\ignorespaces}
\begin{document}
\title{Chasing the Intruder: A Reinforcement Learning Approach for Tracking Intruder Drones}
\author{Shivam Kainth, Subham Sahoo, Rajtilak Pal, Shashi Shekhar Jha} \affiliation{%
\institution{Indian Institute of Technology Ropar}
\country{India}
}

\email{{shivam.20csz0006,2020csb1317,2022aim1010,shashi}@iitrpr.ac.in}

    

    




\begin{abstract}
Drones are becoming versatile in a myriad of applications. This has led to the use of drones for spying and intruding into the restricted or private air spaces. Such foul use of drone technology is dangerous for the safety and security of many critical infrastructures. In addition, due to the varied low-cost design and agility of the drones, it is a challenging task to identify and track them using the conventional radar systems. In this paper, we propose a reinforcement learning based approach for identifying and tracking any intruder drone using a chaser drone. Our proposed solution uses computer vision techniques interleaved with the policy learning framework of reinforcement learning to learn a control policy for chasing the intruder drone. The whole system has been implemented using ROS and Gazebo along with the Ardupilot based flight controller. The results show that the reinforcement learning based policy converges to identify and track the intruder drone. Further, the learnt policy is robust with respect to the change in speed or orientation of the intruder drone. 
\end{abstract}




\maketitle

\section{Introduction}
In recent years, multiple use cases for Drones (unmanned aerial vehicles) have emerged. From delivery of medicines and vaccines to remote locations, to security-critical applications such as border security and surveillance, along with search and rescue operations during natural disasters \cite{erdelj2016uav}, drones have proven their utility and effectiveness as an affordable and versatile technology. Drones have also been extensively used in object detection and tracking applications, wildlife protection, road and railroad surveying, and critical infrastructure security. 
Due to the versatile nature of drones, they are now being exploited for nefarious purposes as well e.g. spying or intruding into private or protected spaces, smuggling of contraband and weapons, etc. Currently, very few works deal with the use of pursuit-evasion techniques for detecting and timely tackling of intrusions caused by such drones. The problem of intruder drones becomes more challenging as they usually remain undetected by the conventional radar system due to their varied miniature designs \& configurations and their agility to fly at very low altitudes with minimal noise and high speeds. Moreover, by the time an intruder drone gets spotted, and a response is put in action to capture or neutralize it, the intruder drone vanishes from the restricted perimeter within moments. Hence, the number of intrusions caused by such drones is rising, and security agencies are grappling with putting adequate measures for detecting intruder drones in place. In this work, we consider the problem of tracking and chasing an intruder drone spotted in restricted airspace. We employ computer vision techniques interleaved within the Reinforcement Learning (RL) framework to detect and track the intruder. 

The RL framework is best suited for learning optimal control policies in dynamic environments. RL allows an agent (such as a drone) to learn from the previous experience gathered by repeated interactions with the environment to complete the intended objective. For our considered drone-based pursuit-evasion task, the environment is dynamic as neither the trajectory nor the intended target of the intruder drone is known apriori. Hence, the chaser drone (RL agent) needs to infer the expected trajectory of the intruder from real-time sensing using an onboard camera. Further, the control dynamics of the chaser drone has to be generated in such a manner that it continuously follows the intruder by optimizing its speed and orientation. It is evident that a conventional control is challenging to cope with the dynamic nature of this tracking task, and hence a learning-based framework would be highly suitable to train the chaser drone.  

In this paper, we formulate the problem of tracking and chasing the intruder drone as a Markov Decision Process (MDP). We use the information captured by the chaser drone's camera to construct the system's current state. The central objective of the MDP problem is to maintain the intruder drone at the center of the chaser's camera's Field of View (FoV). We propose a reward function that optimizes two distinct sub-objectives: the chaser drone's alignment and chasing speed to follow the intruder drone. Finally, we use a Deep-RL algorithm that has shown superior performance in continuous control tasks called Deep Deterministic Policy Gradients (DDPG) method \cite{lillicrap2015continuous} to learn the policy for the chaser drone.Our policy training and testing inherently considers the real-world dynamics of the actual environments. 


\section{Related Work}
\label{sec:lit}
There have been various works on detection and tracking using drone camera images in the literature.
 In \cite{vanegas_uav_2017-1}, the problem of flying a drone to track a ground robot is formulated where the authors solve the problem using Partially Observable MDPs. The paper assumes a discrete environment where a drone follows the ground object using classical MDPs. The results clearly show that application to the real world is challenging for such sample-based discrete approaches.
 Another approach for drone navigation to the desired location is discussed in \cite{feng_infrared_2021}, where the authors have used infrared sensors to detect a target location and hover over it. They successfully demonstrate the learned policy in an indoor environment. Although, the choice of sensor limits the outdoor applications of this approach.
 In \cite{chen_autonomous_2020-1}, the authors have localized  a radio frequency mobile target. In this work, a target is assumed to be emitting omnidirectional radio frequency signals. The drone swarms are equipped with received signal strength (RSS) sensors and minimize the distance to the ground target based on the signal strength. 

In \cite{pham_autonomous_2018}, the authors have discussed a method of drone navigation by dividing the terrain into grids and assigning Q-values to all states. This approach can only be applied in simple, indoor, and known environments and is inefficient in real-world situations. The paper deals with a small size state space ($\approx 25$), which is significantly less for any real-world use case. 
 In \cite{akhloufi_drones_2019}, the authors have proposed an Action-Decision network (AD-Net) based approach to track drones in an image frame. The AD-net helps is deciding where to move the boundary box for the next time step. 
 In \cite{fernandes2019drone},  the authors have proposed a system capable of identifying various objects in the sky. It can distinguish between birds, clouds, and drones. Hence, it can detect false positives and provide accurate detection of drones.  In \cite{zhang_eye_2019}, the  author has proposed a novel method of detection of various objects from frames captured by drone cameras using the Intersection-over-Union (IoU) metric. In \cite{zhu2020multi}, the authors have proposed a method for using a multi-level Siamese feature extraction module for efficient detection and tracking of drones in video frames. In\cite{hong2021multitarget}, to achieve real-time tracking of drone targets, the authors have used YOLOv3 for detection and DeepSORT for continuous tracking of targets in video frames.

 As noted from the above discussions, although there are various works for identifying and detecting drones using camera images, there has not been any significant work for continuously tracking an intruder drone using reinforcement learning-based controls. In this paper, we address this gap in the literature with our proposed approach.

\section{System Description}
\label{sec:system}

In this paper, we consider one chaser drone, denoted by $D$ whose objective is to track and follow an intruder drone denoted by $E$. Further, we assume that the chaser drone $D$ is equipped with a single monocular camera mounted in front to keep the intruder drone $E$ in its FoV. Further, the intended trajectory of $E$ and its moving velocity are entirely unknown to $D$. 

We consider that $E$ is being followed or tracked by $D$, if $E$ is always present in $D$'s FoV with some pre-specified configuration. Hence, we can describe drone tracking as:
\begin{equation}\label{eq_track}
    \parallel loc_{D}(x,y)-loc_{E}(x,y) \parallel \leq{min(W,H)} 
\end{equation}
where $loc_{D}(x,y)$ is the coordinates of the center of $D$'s FoV and $loc_{E}(x,y)$ is the coordinate of $E$ in $D$'s FoV. $W,H$ are the width and height in pixel lengths of $D$'s FoV respectively. 

The size of $E$ in $D$'s FoV denoted as $\text { Size}\left(E, \text{FoV}\right)$ infers a noisy estimate about the distance between $D$ and $E$, $\text { Dist }{(E, D)}$. We use this noisy estimation for providing location awareness to $D$ w.r.t. $E$. 

\begin{equation}
   \text { Dist }{(E, D)} \propto {1}/{\text { Size}\left(E, \text{FoV}\right)}
\end{equation}
The overall objective is to decrease the distance between the chaser drone $D$ and the intruder drone $E$ while keeping it at the center of FoV of $D$. This can be given as:
\begin{equation}
\min \sum_{t=0}^{\infty}\{\parallel loc_{D}(x,y)-loc_{E}(x,y) \parallel+\text{ Dist}(E, D)\}
\end{equation}


\section{Proposed Methodology}
\label{sec:proposed}
In this paper, we propose to learn an RL-based control policy for autonomously tracking and chasing an intruder drone by a chaser drone. We further designed a computer vision pipeline to detect $E$ from $D$'s FoV. For detection, we used the YOLOv5 \cite{glenn_jocher_2020_4154370} network for the detection task. Post detection, the identified position of $E$ from $D$'s FoV is filtered and provided as an input to the RL algorithm. Next, we describe our computer vision-based detection module for detecting intruder drone from the FoV of the drone's camera.

\subsection{Intruder Detection}
The chaser drone $D$ captures raw frames using its camera. For detecting  $E$, we considered employing the well-established object detection deep learning framework popularly known as the YOLOv5 network. The detector identifies the intruder drone $E$ and provides us with its bounding box coordinates, $<X_\text{low},Y_\text{low},X_\text{high},Y_\text{high}>$. We trained the YOLOv5 network using a dataset designed by manually annotating 5000 images of $E$ captured by $D$'s camera. Considering the different orientations during drone maneuvers, the images were transformed by applying rotation and scaling techniques, thus increasing the training dataset to 20,000 images. The resultant model is found to be very robust that can detect drones in 97\% of the frames. Moreover, the detection model is scalable and can be extended to real-world drone tracking in drone camera images due to the extensive training performed for the same.

\subsection{MDP Formulation}
A RL problem is often framed as a \textit{Markov Decision Process} (MDP) where the state of the environment is fully visible to the agent. However, for this problem, the environment is partially observable and therefore, a limited amount of history information is encoded in the state to make the system Markovian. 
An MDP is defined as a tuple $<\mathcal{S}, \mathcal{A}, \mathcal{P}, \mathcal{R}>$ where $\mathcal{S}$ is the set of states of the environment, $\mathcal{A}$ is the set of actions that are allowed to be taken in the environment, $\mathcal{P}$ is the transition probabilities of the environment and $\mathcal{R}$ is the set of rewards that are provided to the agent based on the outcome of its actions. As can be noted, in this problem the model of the environment $\mathcal{P}$ is unavailable. Hence, drones need to learn a policy by sampling from the environment with repeated interactions.
\begin{description}
     \item[\textbf{States}]:  In the considered framework, the drone has access to the images captured by the mounted camera and its own velocity. The camera image $\operatorname{FoV}(D)$ is communicated using ROS to the intruder module which is processed using YOLOv5, which detects $E$ in  $\operatorname{FoV}(D)$ and returns the pixel coordinates. These pixel coordinates are then used as the state space for the chaser drone $D$.
The overall observation space of $D$ is a total of 8 components represented as $O:<X_\text{low},Y_\text{low},X_\text{high},Y_\text{high},V_{X},V_{Y},V_{Z},Y_{D}>$where $<V_{X},V_{Y},V_{Z}>$ are the velocities of $D$ along $<X,Y,Z>$ axes and $Y_{D}$ represents the current orientation of $D$. Further, we kept five such previous observation tuples to form a single state of the environment at any time point.

\item[\textbf{{Actions}}]:
The action space of the chaser drone is defined as $\mathcal{A}:<V_{X},V_{Y},V_{Z},$ $Y_{D}>$, where the first three values represent velocities in forward, lateral, vertical directions. $Y_{D}$ represents the heading angle of the drone. All the considered parameters of the action tuple are continuous. Further, the range of each component of actions has been clipped in the range of $(-1,1) \ m/s$.

\item[{\textbf{Rewards}}]:
The most crucial aspect of learning a RL policy is to design a well-suited reward model. Rewards encode the objective of the system within the MDP problem.
In our system, we consider two kinds of rewards namely $R_\text {track} $ and $R_{\text{align}}$. This reward model ensures that the chaser drone $D$ must always keep
the intruder $E$ at the center of $\operatorname{FoV}(D)$, as well as reduce the
distance between the intruder $E$ and the chaser $D$, $\text{Dist }({E, D})$.
$R_{\text{align}}$ is calculated as the Euclidean distance between
the center of $\operatorname{FoV}(D)$ i.e. $loc_{D}(x,y)$, \& $loc_{E}(x,y)$ while
$R_{\text {track}}$ is the total perimeter of the bounding box encompassing 
$E$ in $\operatorname{FoV}(D)$, given as:

\begin{equation}
R_{\text {align }}=\left\{\begin{array}{cc}
{[-100,5),} & \text { if } E \in \operatorname{FoV}(D) \\
0, & \text { if } E \notin \operatorname{FoV}(D)
\end{array}\right\}
\end{equation}

\begin{equation}
R_\text{track}=\left\{\begin{array}{cc}
    {[-100,5),} & \text{if} \ \Pi(E,\operatorname{FoV}(D)) > 0 \\
    0 , & \text{if} \  \Pi(E,\operatorname{FoV}(D)) = 0 
    \end{array}\right\}
\end{equation}
\begin{equation}
\Pi(E,\operatorname{FoV}(D))= \left\{\left[X_{\text {high }}-X_{\text {low }}\right]+\left[Y_{\text {high }}-Y_{\text {low }}\right]\right\} \times 2
\end{equation}

A heavy penalty for the chaser drone is defined for the cases when $E$ goes out of sight for an extended duration of time. Hence, if for 50 steps, $D$ is unable to spot $E$ in $\operatorname{FoV}(D)$, the episode ends early with a penalty of -200. Another case is when $D$ moves very close to $E$ thus increasing the chances of a collision, it incurs a penalty of -100 however the episode continues.

$$\text{Penalty, }P=\left\{\begin{array}{l}
    -100, \text { if } \text{ Dist }({E, D}) \leq 0.5 \mathrm{~m} \\
    -200, \text { if } E \notin \operatorname{FoV}(D) \forall t \in(0,50)
    \end{array}\right\}$$

The total reward ($R$) of the system is given as:

\begin{equation}
R=\left(R_{\text {track }}+R_{\text {align }}+P\right)
\end{equation}
\end{description}

\subsection{Learning the Chaser Policy}
\begin{figure}
\includegraphics[width=\textwidth/2]{"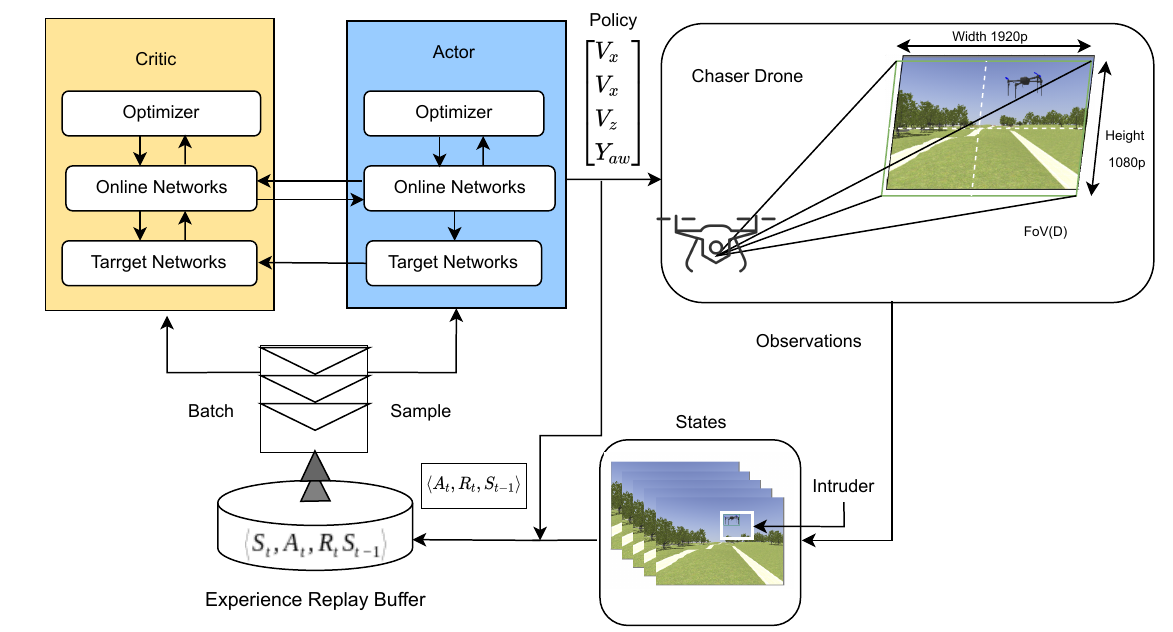"}
\caption{The proposed DDPG based model for learning continuous controls of the chaser drone. }
    \label{fig:ddpgarc}
\end{figure}

Each action in our action space $\mathcal{A}$ is a continuous action that motivates the use of \textit{Deterministic Policy Gradients} \cite{silver2014deterministic}. DDPG uses an actor-critic architecture consisting of two main deep neural networks: the Actor Network and the Critic Network. The overall architecture of the DDPG method is depicted in Figure \ref{fig:ddpgarc}. The Actor-Network $\mu\left(s|\theta^\mu \right)$ is used to generate the actions to be executed by the chaser drone in the environment while the Critic Network $Q \left(s,a|\theta^Q \right)$ is used to assess the viability of the actions generated by the actor. This algorithm fits perfectly with our objective as we have to precisely control $D$ using translatory actions and heading angle. 
There are two more networks in DDPG: Target Actor Network $\mu'\left(s|\theta^{\mu'} \right)$ and Target Critic Network $Q' \left(s,a|\theta^{Q'} \right)$ which are architecturally identical to the Actor and Critic Networks respectively. Directly implementing learning update with neural networks makes it unstable in many environments.

DDPG keeps a replay buffer $\Re$ which stores a finite number of samples from the environment that the drone has collected in the recent interaction cycles in the form of $\left(s_t, a_t, r_{t+1}, s_{t+1} \right)$ where $s_t$ is the current state of the drone, $a_t$ is the action that was taken in $s_t$, $r_{t+1}$ is the reward observed after taking $a_t$ in $s_t$ and $s_{t+1}$ is the next state that the drone landed up in. The replay buffer addresses the issues of sample inefficiency and makes the updates more productive.

In each step of the episode, a random mini-batch of samples is taken uniformly at random and the actor and critic networks are updated. The actor parameters are updated in the direction of the gradient of the performance objective $\mathcal{J}$:

\begin{equation}
    \nabla_{\theta^\mu}\mathcal{J} \approx \mathbb{E}_{s_t \sim \rho^\beta} \left[ \nabla_a Q \left( s,a |\theta^Q \right)|_{s=s_t, a = \mu(s_t)} \nabla_{\theta^\mu} \mu \left(s|\theta^Q \right)|_{s=s_t} \right]
\end{equation}
where the expectation is taken over states $s_t$ coming from a discounted state visitation distribution for a stochastic behavior policy $\beta$. The critic is updated by minimizing the expected loss between the critic's value and the target generated from the target critic network which is given as:
\begin{equation}
    L \approx \mathbb{E}_{s \sim \rho^\beta} \left[\left(r_{t+1} + \gamma Q'\left(s_{t+1}, \mu'\left(s_{t+1}| \theta^{\mu'} \right)|\theta^{Q'} \right)  - Q \left(s_t, a_t | \theta^Q \right)\right)^2\right] 
\end{equation}

The parameters of the target networks are updated by having them slowly track their original networks: $\theta' \leftarrow \tau\theta + \left(1- \tau\right)\theta'$ where $\tau << 1$. 

\section{Experimentation}
\label{sec:experiment}
\begin{figure}
\includegraphics[width=\textwidth/2]{"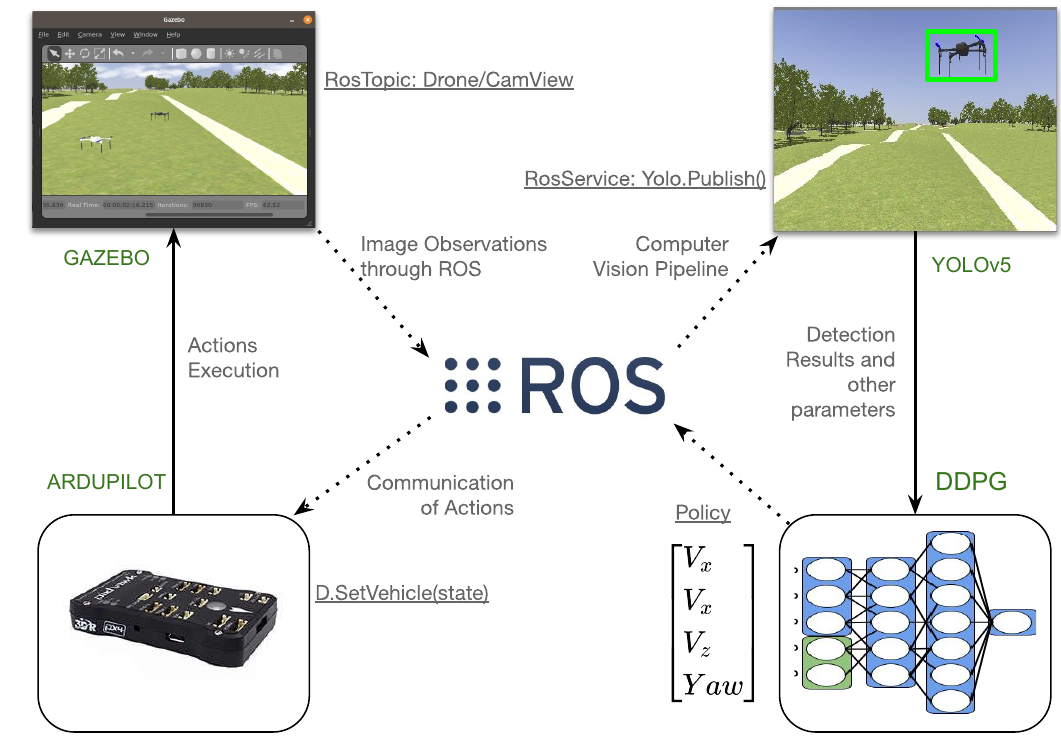"}
\caption{The overall system architecture as implemented using Gazebo and ROS. }
    \label{fig:sysarc}
\end{figure}

For training and evaluating the proposed approach, we implemented the whole system of chaser and intruder drones using Gazebo\cite{1389727} and ROS\cite{ROS}. Gazebo is a 3D simulator with a high-performance physics engine, which is capable of simulating real-world situations and interactions using various sensors such as cameras, LiDARs, and GPS. 
ROS is a well-known open-source middleware for implementing robot functions. ROS uses a subscriber-publisher model with a set of libraries and tools to provide communication among multiple modules with a robotic system. 
As a drone controller, we have used Ardupilot\cite{ardupilot_ardupilot_nodate} which is a open-source flight controller used in various aerial vehicles such as multi-copters, helicopters, fixed Wing drones, and rovers. 
In our implementation, Gazebo provides a 3D simulation platform, ROS is used as the communication framework between the chaser drone and the Gazebo environment and Ardupilot is  used to provide flight maneuvers for the chaser drone based on the learned control using DDPG model. The overall system implementation view is depicted in Figure \ref{fig:sysarc}. As can be seen in the figure, ROS is at the core of whole implementation providing middleware support using its publish-subscribe framework. The multiple ROS topics are depicted in the figure that performs specific functions such as drone image capturing, drone detection, training and translation of actions to the drone.

\subsection{Training the Chaser Drone}
The environment is simulated in Gazebo with varying background scenes. The simulation has two Iris quadcopter drones set in rough terrain with mountainous features, forests, and highways with multiple vehicles moving randomly in the background. 
During each training episode, the starting locations of intruder and chaser drones are randomly selected within the range of $(-5,5)$ for each dimension in the $<X,Y,Z>$ direction. This ensures robust training and accommodates multiple scenarios. 
The intruder's velocity is varied from episode to episode  ranging between 1 to 5 m/s. Further, the intruder drone $E$ is also made capable to increase its speed during evaluation if the chaser drone $D$ reaches very close to it. In this way, $E$ can evade $D$ for a long time. 

The complete training runs of the DDPG model are performed  on a DELL Server with Intel Xeon Processor with NVIDIA Quadro RTX A4000 8 GB graphics card and 64 GB RAM. Gazebo simulation is implemented on an ASUS system with AMD 5800H processor, with 16 GB RAM and NVIDIA RTX 3060 6GB graphics card. Python API calls via Flask Framework are used to communicate between the DDPG training and simulation processes. This distributed implementation helps in streamlining the processes and provides a controller-responder architecture that can be used to scale the training to a large number of clusters if required. 
The training is performed over a total of 6000 episodes,  and the experiment is repeated 20 times to validate the results. Further, we constructed an exploration policy $\beta$ by adding noise, sampled from a Ornstein-Uhlenbeck (OU) process \cite{uhlenbeck1930theory} $\mathcal{N}$, to our actor policy $\mu$, given by 

\begin{equation}
    \beta \left(s_t \right) = \mu \left(s_t|\theta_t^\mu \right) + \mathcal{N}
\end{equation}
Episodes are truncated after the agent has moved 750 time-steps, or $E$ is out-of-sight for continuous 50 time-steps. The values of the various hyper-parameters are listed in Table \ref{table:hyperparm}.
\begin{table}[]
\begin{tabular}{|l|r|}
\hline
\textbf{Hyper-parameter}        & \textbf{Value} \\ \hline
Discount Factor $\gamma$       & 0.99           \\ \hline
Mini-batch size                 & 128            \\ \hline
Actor Learning Rate            & 0.001          \\ \hline
Critic Learning Rate           & 0.001          \\ \hline
Replay Buffer Size             & 100000         \\ \hline
Target Update Parameter $\tau$ & 0.001          \\ \hline
\end{tabular}

\caption{Hyper-parameters for DDPG model used for training.}
\label{table:hyperparm}
\end{table}

\subsection{Deep Network Architectures}
In this subsection, we describe the implementation details of the deep neural network architectures used for the DDPG model. The Actor Network, $\mu\left(s|\theta^\mu \right)$, consists of three fully connected hidden layers of 256 units each. The input is a state tensor of shape $\left( 8, 5 \right)$. The output of the network is a vector of four elements representing each action element in $\mathcal{A}$. The Critic Network, $Q (s,a|\theta^Q)$, consists of two input heads. The first input head is of the state tensor of shape $\left(8, 5 \right)$, and the second input head is of the action vector of four elements. The two heads are followed by a fully connected layer, each with 192 and 64 units, respectively, and then their resulting tensors are concatenated together. There are two more fully connected hidden layers of 256 units each, followed by a final output layer of 1 unit representing the Q value for the input state and action. Each of the fully connected layers is followed by a Batch Normalization layer to minimize covariance shift during training.

\subsection{Performance Metrics}
During training, we tracked our DDPG model's progress using some metrics, even though it is challenging to accurately evaluate output policy. We describe some of the metrics that helped us keep track of the training process:

\begin{description}
\item[{\textbf{Total Reward}} ]
The total reward is the measure of the sum of all rewards the chaser drone has received over time due to its action executions in the environment. The more reward the agent accumulates in an episode, the better is the policy. However, we cannot be sure about the robustness of the policy from this metric.




\item[{\textbf{Absolute Value Error}}] 
After every episode, we calculate the difference between the actual return received from that episode and the Q-value predicted at that state for every state-action pair visited. This helps us understand whether the agent knows what it's doing or is completely lost.
\end{description}

Next, we present the results gathered during the training and further evaluation of the trained policy of the chaser drone for various test scenarios.
\begin{figure}
    \centering
    \includegraphics[width=\textwidth/2]{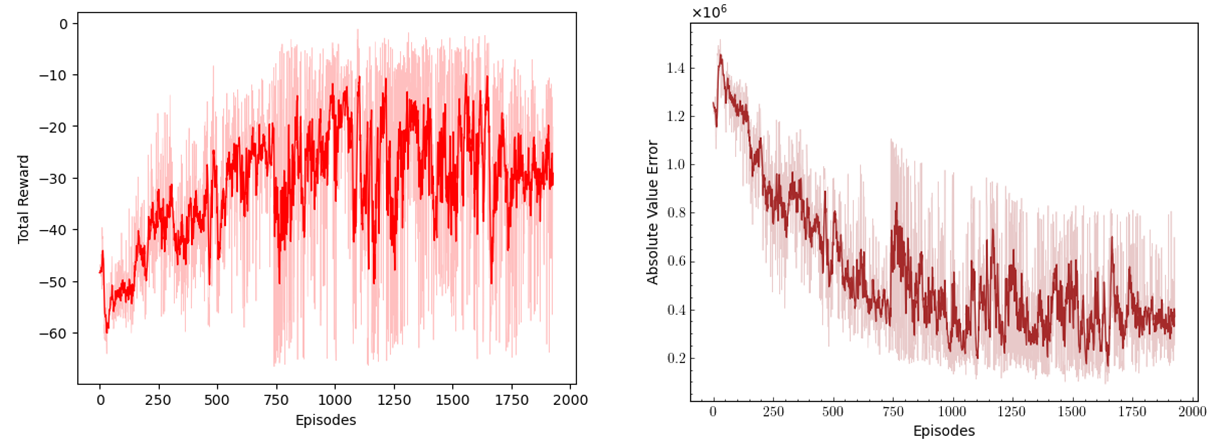}
    \caption{(a) Total reward per episode and (b) absolute value error, during training episodes of the chaser drone.}
    \label{fig:trainmetric}
\end{figure}

\section{Results}
\label{sec:result}
In this section, the performance of the framework is evaluated on various metrics as described in Section \ref{sec:experiment}. 
While testing the system, the policy is not updated, and only the trained weights are used to generate control actions by the chaser drone. 
A total of 500 evaluation episodes with random starting locations to both $D$ and $E$ are used to evaluate the effectiveness of the learned control policy for tracking the intruder by the chaser drone.   


\begin{figure}
    \includegraphics[width=\textwidth/2]{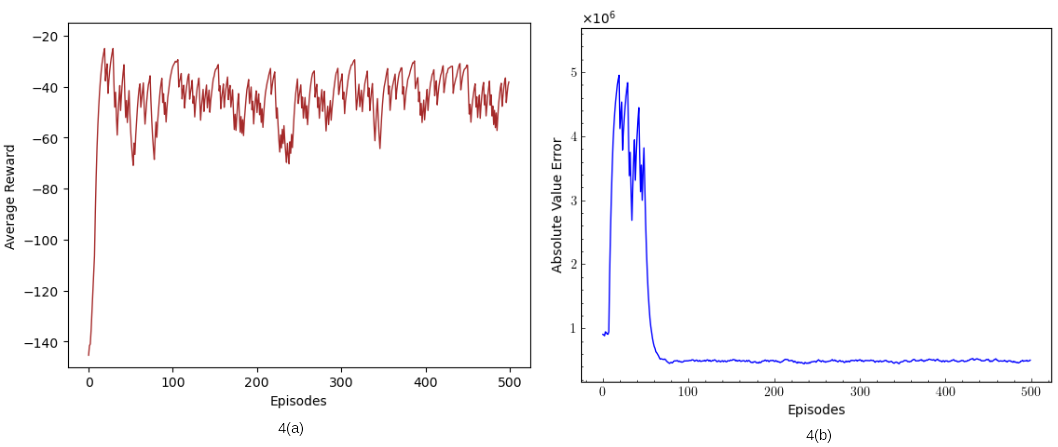}
    \caption{(a) Total reward per episode and (b) absolute value error, during testing when the trained policy was only used without any policy update.}
    \label{fig:evalmetrics}
\end{figure}
\begin{figure}
\includegraphics[width=\textwidth/2]{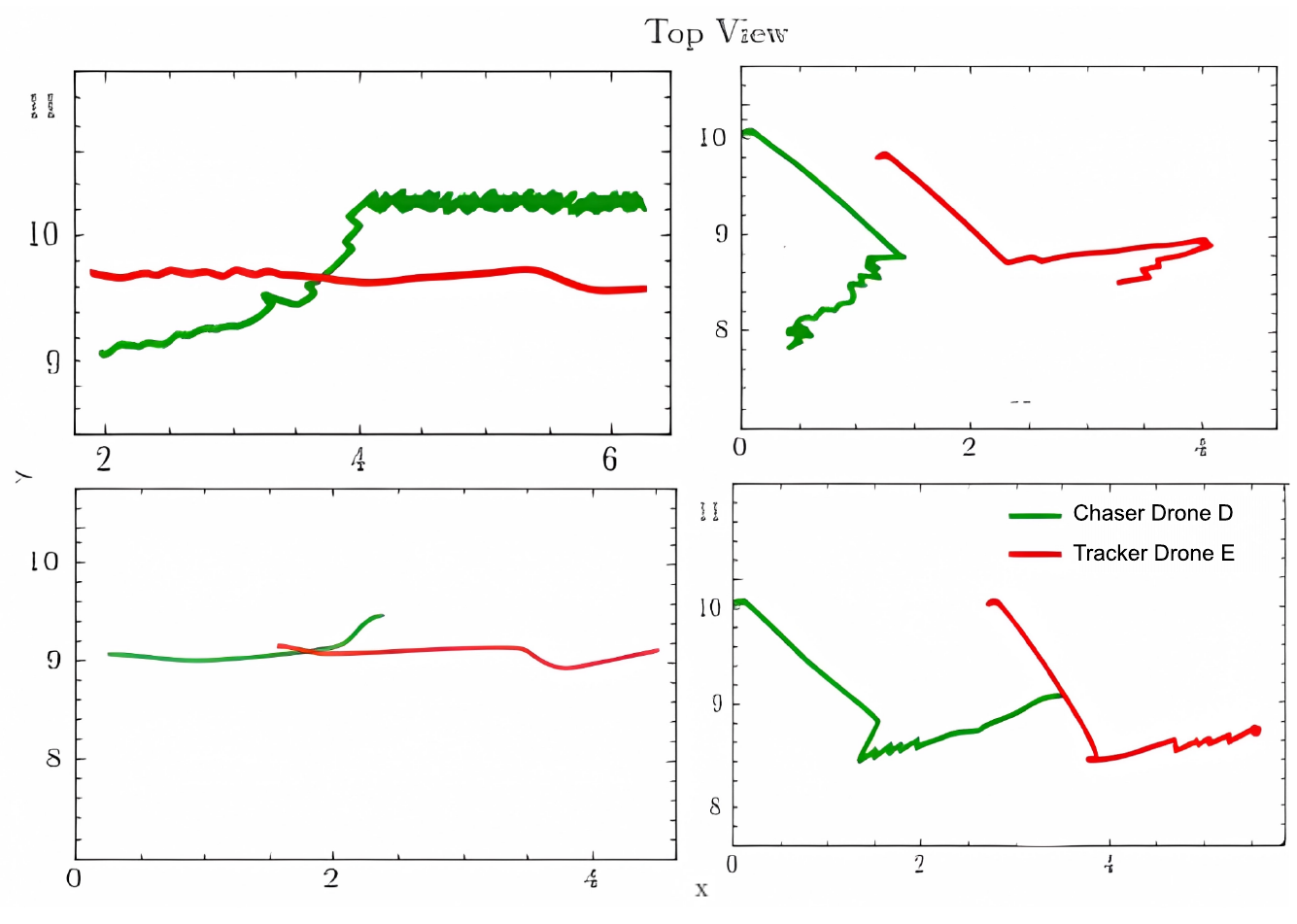}
\caption{Trajectories of the chaser and intruder drones during four different test episodes. The green color shows the chaser drone $D$ trajectory while the red one is for intruder drone $E$.}
    \label{fig:em}
\end{figure}
\begin{figure}
    \centering
    \includegraphics[scale=0.5]{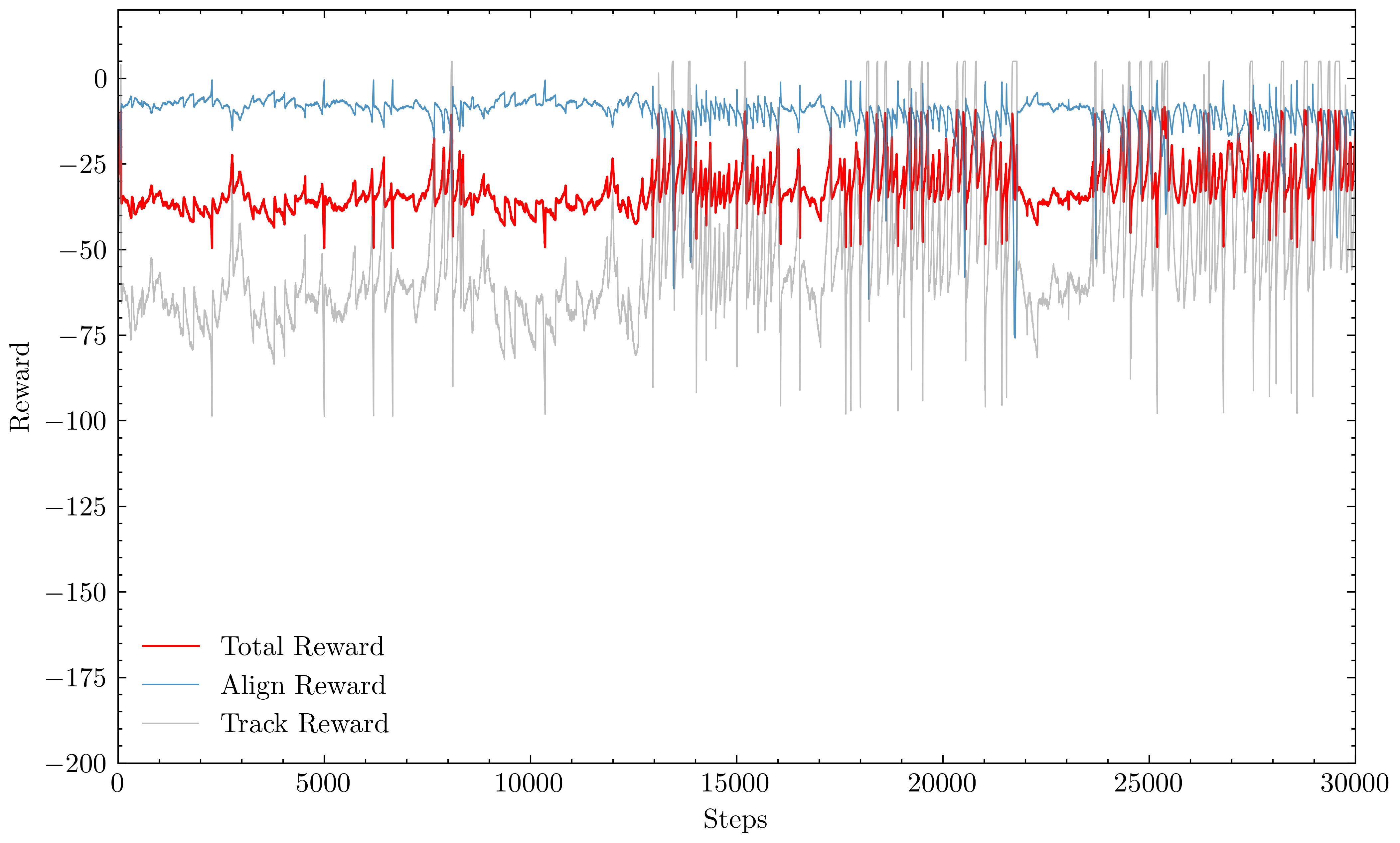}
    \caption{ Align and Track rewards per step for the long endurance chase during testing when the episode ran continuously for four hours in Gazebo.}
    \label{fig:endurance}
\end{figure}

\subsection{Performance during Training}
We ran training simulations in Gazebo and ROS, where the proposed DDPG based model is trained to learn the chaser control policy over multiple episodes. We ran the training for more than 2000 episodes until the evaluation metrics started showing a steady-state performance.

Figure \ref{fig:trainmetric}(a) shows the total reward per episode during training. As can be observed, the episodes have a steady increase in the total reward. The average reward increases slowly initially, directly correlating with the episode's length. Initially, episode lengths are minimal, but later on, as $D$ can track $E$ for a longer duration, total reward starts improving steadily. There is a constant exploration parameter coming from OU noise due to which $D$ keeps trying to find alternative strategies for tracking $E$. The steady improvement in the reward graph shows that the chaser drones has learned to track $D$ resulting in higher reward accumulation in the later episodes. 
Further, with respect to $R_{\text {align }}$ and $R_{\text {track }}$'s influence on total reward, $R$, we observed that in the beginning, there is very less contribution of $R_{\text {track }}$ to total reward $R$. However, as the episodes progress, $R_{\text {track }}$ keeps on improving, as $D$ learns to follow $E$ and this contributes to increase in $R$. At times, $D$ tries to maximize $R_{\text{align}}$ at the expense of $R_{\text {track }}$, but as episode count increase, $R_{\text {track }}$ starts improving with a slight reduction in $R_{\text {align }}$. As evident from the total reward plot, $D$ is able to align properly with $E$ and in later episodes, $D$ focuses more on reducing it's distance to $E$ for continuous tracking. This highlights the effectiveness of the proposed approach to learn a control policy for chasing an intruder drone.

Figure \ref{fig:trainmetric}(b) shows the absolute value error during training as gathered from the critic network loss function of the DDPG model. As can be seen, the absolute value error shows a downward trend becoming asymptotic towards the $2000^\text{th}$ episode. This further shows that the DDPG model is converging as the model is trained with our modeled reward function.

\subsection{Performance of the Control Policy during Evaluation}
As described in Section \ref{sec:experiment},  we ran an evaluation of the trained policy for 500 episodes with different start configurations. Figure \ref{fig:evalmetrics}(a), shows the average reward per episode received by the chaser drone $D$ while tracking $E$. From the graph, it can be observed that $D$ is getting consistently good rewards and can track $E$ in 457 out of 500 episodes. Figure \ref{fig:evalmetrics}(b) shows the absolute value error for $D$ during evaluation episodes. Absolute value error shows a sharp reduction in value during initial episodes, and after that values are relatively stable. These plots clearly indicate the stable system performance of the chaser drone in the identification and tracking tasks of the intruder.

Further, we depict the trajectories of the chaser and intruder drones on four sample episodes out of the 500 evaluation episodes. In Figure \ref{fig:em}(a), it has been observed that $D$ can mimic the initial descent maneuver of $E$. $D$ then  continues adjusting while mainly keeping to the left of $E$. 
Figure \ref{fig:em}(b) shows that the Penalty $P$ component of the Reward $R$ can prevent $D$'s collisions with $E$. $D$  initially descends slowly while reducing distance, but instead of directly colliding with $E$, $D$ maintains a safe distance to $E$.
In Figure \ref{fig:em}(c), it is observed that the fluctuations in the trajectory of $E$ are not causing large disturbances in the trajectory of $D$. If the trajectory of $E$ is relatively straight, $D$ is taking smaller steps as signified by the dense points near the end of the episode. 
In Figure \ref{fig:em}(d), the initial actions of $D$ is always to move down, it is because $D$ quickly tries to decrease $R_{\text{align}}$ error before minimizing the $R_{\text{track}}$ error. This kind of behavior seems logical as, $R_{\text{track}}$ starts improving only when $D$ and $E$ are aligned properly.

We also ran a long endurance test for the chaser drone wherein a single episode was continuously kept running for four hours ($\sim$ 30000 time-steps). Figure \ref{fig:endurance} shows the performance on continuous run of episode for 7000 time-steps. During the initial steps, reward is very less, due to random initialization of both $E$'s velocity and location. Slowly, $D$ starts to improve on $R_{\text{align}}$ and later on after getting a good enough $R_{\text{align}}$ reward, $D$ starts to improve on $R_{\text{track}}$. With thirty minutes of flight time of Commercial Drones on average, the policy is able to continuously  run until the intruder runs out of battery power.

\section{Conclusions and Future Work}
\label{sec:conclude}
In this paper, we used computer vision techniques interleaved within a reinforcement learning framework to detect and continuously track an intruder drone using a chaser drone. Our chaser drone is able to rely on object detection techniques to detect and accurately chase the intruder. Our proposed reward model accurately captures the objectives of aligning the chaser drone and tracking the intruder continuously. The proposed approach has been implemented in using Gazebo and ROS along with Ardupilot as the flight controller. Such an implementation makes it feasible to deploy our system in the real world. The results gathered by evaluating the proposed approach using different configurations in Gazebo simulation validate the effectiveness of our proposed RL based approach. 

Our future work includes deploying a swarm of chasers to pursue a intruder and ultimately neutralize it. Further, we plan to develop hierarchical policies for a swarm of drones to take off, pursue and revert to the recharging station autonomously for intelligent pursuit and effective protection of restricted airspaces.  



\nocite{*} 


\end{document}